\documentclass{article} % For LaTeX2e
\usepackage{iclr2025_conference,times}
\usepackage{graphicx} 
\usepackage{booktabs} 

\usepackage{amsmath,amsfonts,bm}

\def\eqref#1{equation~\ref{#1}}
\def\1{\bm{1}}

\DeclareMathAlphabet{\mathsfit}{\encodingdefault}{\sfdefault}{m}{sl}
\SetMathAlphabet{\mathsfit}{bold}{\encodingdefault}{\sfdefault}{bx}{n}

\usepackage{hyperref}
\usepackage{url}

\title{Cracking the Code of Action: A Generative Approach to Affordances for Reinforcement Learning}

\author{Lynn Cherif \thanks{Equal contribution} \ $^{\text{1,2}}$, Flemming Kondrup \footnotemark[1] \ $^{\text{1,2}}$, David Venuto $^{\text{1,2}}$, 
\AND
Ankit Anand $^{\text{1,2,3}}$, Doina Precup $^{\text{1,2,3}}$, Khimya Khetarpal $^{\text{2,3}}$ \\ \\
$^{\text{1}}$ McGill University, $^{\text{2}}$ Mila, $^{\text{3}}$ Google DeepMind\\ \\
\texttt{lynn.cherif@mail.mcgill.ca} \\ 
}

\iclrfinalcopy % Uncomment for camera-ready version, but NOT for submission.
\begin{document}

\maketitle

\begin{abstract}
Agents that can autonomously navigate the web through a graphical user interface (GUI) using a unified action space (e.g., mouse and keyboard actions) can require very large amounts of domain-specific expert demonstrations to achieve good performance. Low sample efficiency is often exacerbated in sparse-reward and large-action-space environments, such as a web GUI, where only a few actions are relevant in any given situation. In this work, we consider the low-data regime, with limited or no access to expert behavior. To enable sample-efficient learning, we explore the effect of constraining the action space through \textit{intent-based affordances} -- i.e., considering in any situation only the subset of actions that achieve a desired outcome. We propose \textbf{Code as Generative Affordances} $(\textbf{\texttt{CoGA}})$, a method that leverages pre-trained vision-language models (VLMs) to generate code that determines affordable actions through implicit intent-completion functions and using a fully-automated program generation and verification pipeline. These programs are then used in-the-loop of a reinforcement learning agent to return a set of affordances given a pixel observation. By greatly reducing the number of actions that an agent must consider, we demonstrate on a wide range of tasks in the MiniWob++ benchmark that: \textbf{1)} $\texttt{CoGA}$ is orders of magnitude more sample efficient than its RL agent, \textbf{2)} $\texttt{CoGA}$'s programs can generalize within a family of tasks, and \textbf{3)} $\texttt{CoGA}$ performs better or on par compared with behavior cloning when a small number of expert demonstrations is available. 
\end{abstract}

\begin{figure*}[ht]
\centering
\raisebox{-.5\height}{%
  \begin{minipage}[c]{0.4\linewidth}
    \includegraphics[width=\linewidth]{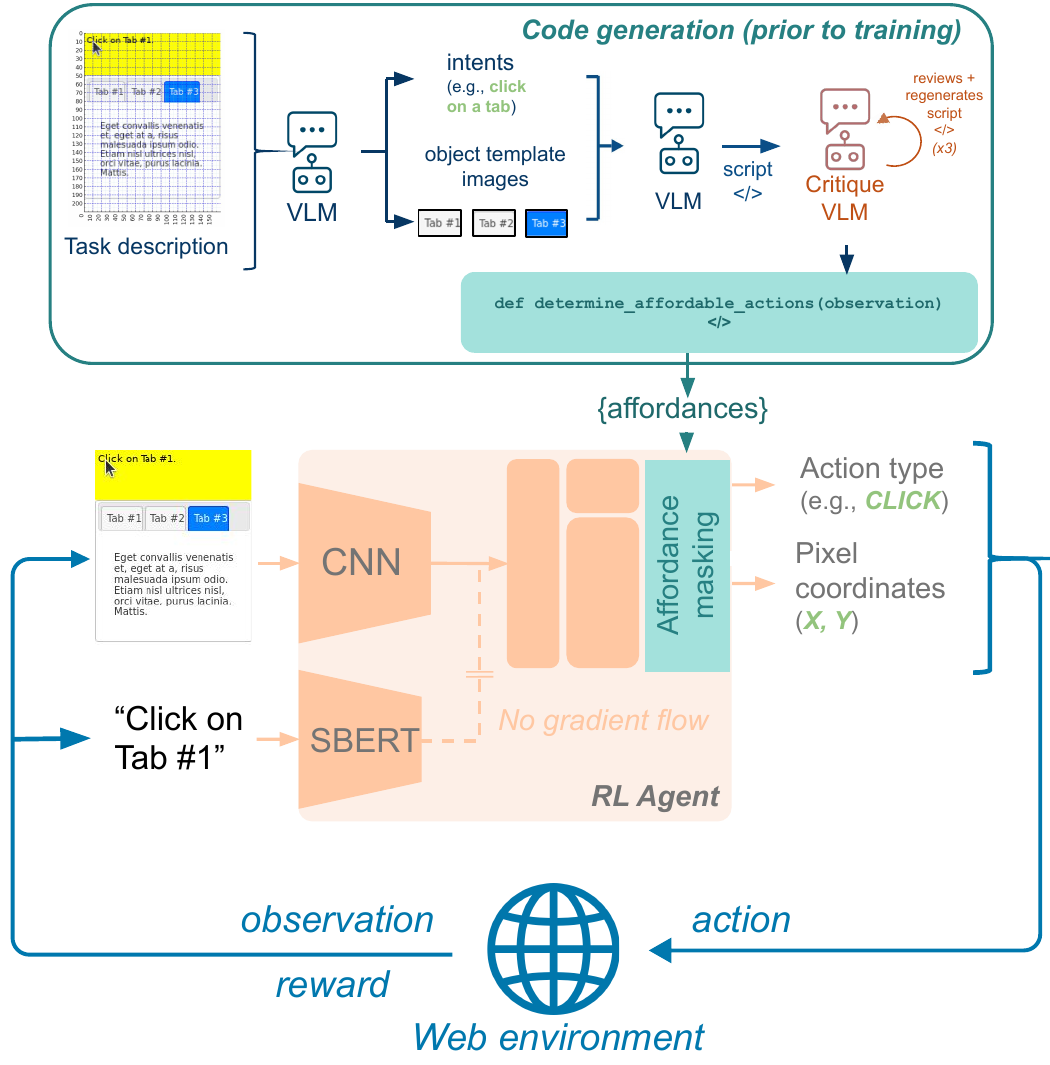}
  \end{minipage}%
}
\quad
\raisebox{-.5\height}{%
  \begin{minipage}[c]{0.5\linewidth}
    \includegraphics[width=\linewidth]{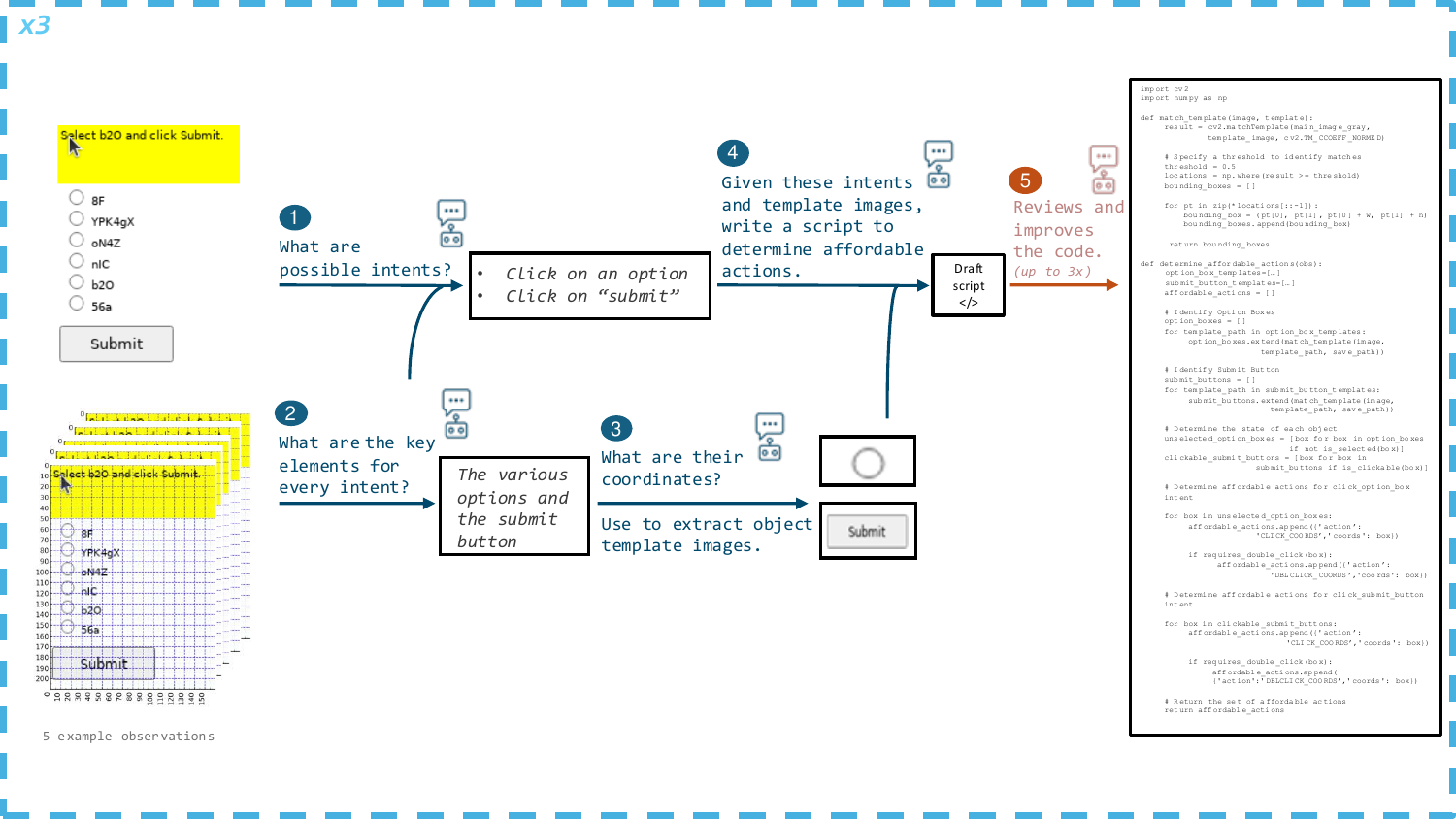}
  \end{minipage}%
}
\caption{\textbf{Left:} Overview of our method, \texttt{CoGA}. The VLM processes available task descriptions and example observations to extract relevant intents (e.g., ``\textit{click on a tab}") and object template images (e.g., every tab), which are then used to generate code that returns a set of affordable actions given an observation. The code is validated and improved by a critique VLM. The set of affordances are then used to mask the action space of the RL agent. \textbf{Right:} Prompting pipeline to generate affordance scripts that return the set of affordable actions.}
\label{fig:coga}
\vskip -0.2in
\end{figure*}

\section{Introduction}
Reinforcement learning (RL) is a powerful paradigm to train agents for sequential decision-making by interacting with an environment. In environments where data collection and human annotation is time-consuming and costly, the sample efficiency of an agent is critical. Despite its great potential and success in multiple domains like Go \citep{silver2016mastering} and Chess \citep{silver2017masteringchessshogiselfplay}, RL algorithms can suffer from significant challenges in being sample efficient. In real-world environments with sparse reward and large action spaces where only a small subset of actions are relevant in a given situation (e.g., GUI-based web navigation, recommendation systems), this issue is exacerbated. 

To address this challenge, a popular approach is to leverage expert trajectories with behavior cloning (BC) (e.g., \cite{shaw2023pixels}). State-of-the-art methods require thousands to millions of such demonstrations. However, this comes with major limitations including computational costs and the burden of gathering domain-specific expert demonstrations. Moreover, BC suffers from an imitation gap~\citep{pmlr-v9-ross10a} and rarely surpasses its training data. In contrast, RL has the possibility to gather new data and learn from interaction. Yet, RL methods alone struggle to bridge the gap to expert performance in many tasks, particularly those with large action spaces and sparse rewards. In this regard, we here focus on reducing the complexity of the action space, and by consequence alleviating the sparse reward challenge. Progress towards improving the performance and sample efficiency of RL agents is complimentary to methods such as \citep{shaw2023pixels}, with potential to further enhance the RL fine-tuning component.

In the context of RL, \citet{khetarpal2020can} defined \textit{affordances} \citep{gibson1977theory} as actions that complete intended consequences (i.e., \textit{intents}). Intent-based affordances help prune the action space and guide RL agents towards effective actions. This mitigates naive exploration and reduces sample complexity. To bridge the gap towards or beyond expert performance in the \emph{low-data regime}, we focus on \emph{learning with limited to no access to expert demonstrations} and investigate the use of affordances to improve the sample efficiency of RL.

However, the specification of intents and intent-completion functions is non-trivial and remains an open problem. For instance, hand-designing them can be limited in environments where intents are not obvious and require substantial effort and domain knowledge. We address this challenge by leveraging pre-trained large vision-language models (VLMs) to discover intents and the corresponding actions they afford. 

Given their multimodal reasoning capabilities, pre-trained VLMs are well-suited to enhance RL agents operating with image-based observations. Although querying a VLM directly for affordances based on visual observations \citep{qian2024affordancellm} or making the VLM itself be the agent is possible, it is computationally and financially expensive. We use VLMs to generate functions that return affordable actions through implicit intent-completion, requiring a robust code generation and verification pipeline. While prior work \citep{venuto2024code} focused on high-level tasks like sub-task reward functions, our prompting and verification approach ensures reliable low-level affordance specification. The generated code is used in the RL training and inference loop for pruning the action space, thus improving the agent's sample efficiency.

Our framework, \textbf{Code as Generative Affordances} \textbf{(\texttt{CoGA})}, demonstrates strong sample efficiency and success rates on a series of MiniWob++ \citep{pmlr-v70-shi17a,wgeliu2018reinforcementlearningwebinterfaces} tasks. While we study \texttt{CoGA}'s efficacy in the challenging domain of automated visual web navigation, the core contribution of our work lies in the methodological framework of generating intent-based affordances as code to enhance the sample efficiency of RL agents. Specifically, we show the following claims (Sec. \ref{sample-eff}):

\begin{enumerate} 
    \item \textbf{\texttt{CoGA} is orders of magnitude more sample efficient than its RL agent.}
    \item \textbf{\texttt{CoGA}'s generated affordance scripts can generalize within the same family of tasks.}
    \item \textbf{\texttt{CoGA} performs better or on par compared to its reference BC agent when only a limited number of expert demonstrations are available.}
\end{enumerate}

\section{Background}\paragraph{Reinforcement Learning.} An RL agent learns to interact with an environment, through a sequence of actions, in order to maximize its expected long-term reward~\citep{sutton2018introduction}. This interaction is typically formalized using the framework of Markov Decision Processes (MDPs). A finite MDP is a tuple $M=\langle {\cal S}, {\cal A}, r, P, \gamma \rangle $, where ${\cal S}$ is a finite set of states, ${\cal A}$ is a  finite set of actions, $r: {\cal S} \times {\cal A}\rightarrow \mathbb{R}$ is the reward function, $P:{\cal S} \times {\cal A} \rightarrow \mathrm{Dist}({\cal S})$ is the environment's transition dynamics, mapping state-action pairs to a probability distribution over next states,  $\mathrm{Dist}(\mathcal{S})$, and $\gamma \in (0,1)$ is a discount factor.  At each time step $t$, the agent observes a state $s_t \in {\cal S}$ and takes an action $a_t \in {\cal A}$ drawn from a policy $\pi : {\cal S}  \rightarrow \mathrm{Dist}(\cal A)$.

\paragraph{Q-learning.}
A value function of a policy $\pi$ is defined as the expectation of long-term return (i.e., the cumulative discounted reward) received from a given state and by executing  $\pi$, defined as: $ V^\pi(s) =  \mathbb{E}[\sum_{t=0}^\infty \gamma^t r(s_t,a_t)|s_0=s] \text{ where } a_t \sim \pi(\cdot|s_t), \text{for all } t$. The action-value function is defined as $Q^\pi(s,a) =  r(s,a) + \gamma \sum_{s'} P(s'|s,a) V^{\pi}(s')$. In Q-learning \cite{watkins1992q}, the optimal action-value function $Q^*$ corresponds to the optimal policy $\pi^*$: $Q^* = \max_{\pi} Q^\pi(s,a)$. The optimal policy $\pi^*$ can be obtained by acting greedily with respect to $Q^*$. In complex environments, the optimal Q-value function can be approximated using a neural network, referred to as Deep Q-learning (DQN) \cite{mnih2013playingatarideepreinforcement}.

\paragraph{Vision-Language Models.} VLMs are pre-trained transformers that integrate visual and textual data as input, enabling multi-modal reasoning. These models consist of three core components: a vision module to process images, a text module for language inputs, and a fusion mechanism —often utilizing cross-modal attention— to link visual and textual embeddings. Models like CLIP \citep{radford2021learning} and UNITER \citep{10.1007/978-3-030-58577-8_7} have demonstrated impressive performance in tasks such as scene description and image-text matching. These models leverage contrastive learning or transformer-based architectures to align images and text in a shared embedding space. More recent, larger models such as GPT-4 \citep{openai2023gpt4} showcase exceptional performance across a range of tasks with increasing complexity.

\paragraph{Intents, Intent Completion, and Affordances.} The concept of \textit{intent} refers to the desired outcome associated with an action ~\citep{gibson1977theory}. Intents are abstract representations of goal states, guiding an agent's decision-making. In RL, \textit{affordances} are the state-action pairs that can \textit{complete} these intents, effectively reducing the action space by focusing only on relevant actions in a given state \citep{khetarpal2020can}. Concretely, an \textit{intent-completion function} considers a transition $(s_t,a_t,s_{t+1})$ and set of intents, and predicts the likelihood of the transition to complete the respective intents above a certain threshold. Thus, affordances can be inferred through the intent-completion function. In this work, we posit that a VLM can predict the likelihood of achieving an intended consequence for a given state and an action, which can lead to relevant affordances. Specifically, we leverage the VLM to 1) specify the relevant intents for a task (e.g., ``click tab"), 2) infer implicit intent-completion functions by iteratively building its understanding of the task, and 3) generate affordances as code by implicitly using its inferred intent-completion functions (see Sec. \ref{sec:coga}).

\section{\texttt{CoGA:} Code as Generative Affordances}
\label{sec:coga}

We now present our approach, \texttt{CoGA}, which leverages pre-trained VLMs to generate code for determining affordable actions given an image observation (see \texttt{determine\_affordable\_actions(obs)} in Figure \ref{fig:coga} - right). These generated functions return a set of affordable actions which can be used to prune the action space in RL. This task not only requires high-level reasoning by the VLM, but also correctly inferring and detecting the low-level affordable actions (i.e., affordable action types and pixel coordinates here) in each and every observation through the generated code. 

\subsection{Generating Affordances as Code. } \texttt{CoGA} proposes a modular prompting pipeline (see Figure \ref{fig:coga} - right and Appendix \ref{appendix:promptingpipeline}) which builds on that of \citet{venuto2024code}. \texttt{CoGA} consists of three key components: \textbf{1)} a modular code generation pipeline that first identifies the correct set of intents and relevant objects given a task description and an image observation, and then generates a function that returns the set of affordable actions given an observation, \textbf{ 2)} a verification pipeline that leverages another VLM for critiquing the generated code and improving it accordingly, the final code (to be used in RL) is selected based on ground truth test cases, \textbf{3)} using the generated code resulting from steps (1) and (2) in RL. This pipeline alleviates the need for expensive VLM inference calls during the RL stage. As shown in Figure \ref{fig:coga} (right), inferring the relevant high-level intents in a task (e.g., ``click on an option", ``click submit") is the first step in generating functions to determine affordances. For every intent, the VLM (GPT-4o) is prompted to determine the  relevant objects (e.g., every tab) and their bounding box coordinates. These objects' bounding boxes thereby correspond to the affordable pixel actions and need to be dynamically detected for every observation through the generated script. To do so, we aim to use off-the-shelf object detection methods that do not require additional training, as in \citet{venuto2024code}. However, unlike \citet{venuto2024code}, we resort to template image matching. We found it more robust for detecting complex and granular objects (e.g., cartoon trash cans) than edge and color detection methods used by their generated reward functions. 

\paragraph{Inferring Intents.}The process starts by prompting the VLM to build context about the environment. We first show it a randomly sampled observation, the task description and example instructions of the task given by the environment. We then ask it to identify the salient objects in the  image, followed by the relevant intents for the \textit{task type}. It is important to distinguish between intents and goals, as a goal is related to directly solving the task by completing a given instruction (e.g., ``click on Tab 1"), whereas an intent is related to solving the \textit{type} of task more broadly (e.g., ``click on a tab"). 

\paragraph{Detecting Visual Affordances.} Once the intents are discovered, the VLM is prompted to name the relevant objects to each intent. For every named object, we follow an automated template image extraction process. The VLM is shown a coordinate-system-based gridded image and is required to specify the bounding box coordinates of the respective objects. The prompting pipeline then queries pre-written code to crop and save the objects' image templates using their determined bounding box coordinates. The saved template images are then used in a pre-written template matching script using OpenCV. The template images are derived from 5 randomly sampled observations to maximize generalization across the observation space. Additionally, to avoid discrepancies in color (e.g., a circle is always a circle regardless of its color), we perform template image matching between the grayscaled template images and observation images.   

\paragraph{Determining Affordable Actions.} Once the affordable objects are detected for every intent, the VLM is asked to develop 4 strategies in sequence using chain-of-thought prompting \citep{wei2023chainofthoughtpromptingelicitsreasoning}, before writing the function that returns the affordable actions for a given observation: 1) a step-by-step strategy for determining which intent(s) are relevant for a given observation, 2) a step-by-step strategy for determining which actions are affordable for a given observation and for each intent, 3) a step-by-step strategy for combining the intents and their corresponding set of affordable actions for a given observation, 4) an outline of the script to determine affordances using code comments. Finally, the VLM is asked to write code to implement its strategy for selecting the applicable intents and their corresponding set of affordable actions, for a given observation, by filling in the outline of comments with code. As such, the VLM uses the extracted template images and template matching script as needed to detect the affordable pixel actions (e.g., tabs). The generated code returns the set of affordable actions (both action types and their corresponding pixel actions) for a given observation.  

\subsection{Verification Pipeline}
\label{subsec:verificationpipeline}

For every task, we automatically verify the generated scripts using a) a critique VLM and b) ground truth test cases. The ground truth test cases consist of 5 randomly sampled observations that we manually annotate with a set of affordable actions. Firstly, the scripts are automatically executed on the 5 randomly sampled test observations to check for execution errors. In the case of any thrown execution errors, the critique VLM reviews them and provides improvement feedback similar to \citet{wang2023voyager}. Else, the scripts' precision and recall are calculated with respect to the manually annotated ground truth set of affordances. As well, the critique VLM is shown 2 of the 5 randomly sampled test observations as a reference for assessing the quality of the scripts and provides feedback accordingly. The feedback is reused by the critique VLM to improve and regenerate the code. This process repeats up to 3 times (we have not empirically observed gains beyond this number), unless the critique VLM approves the code earlier. Note that the critique VLM does not have access to the ground truth affordances of the test cases. We log the mean precision and recall over the  5 manual test cases and run the pipeline a maximum of 3 times. We retain the best performing scripts across pipeline runs and critique iterations. 

\subsection{Using the Generated Affordance Script in RL}
The generated affordance script is queried in the training and inference loops of the RL agent to obtain the set of affordable actions. As shown in Figure \ref{fig:coga} (left), the predicted set of affordable actions is used to create a hard mask over unaffordable actions (i.e., the probability of sampling unaffordable actions is 0). Thus, it is important to highlight that the success of \texttt{CoGA} strongly depends on the quality of the generated script, which in turn depends on the accuracy of the object detection method used. If the predicted affordances have low recall, \texttt{CoGA} would fail. In such a case, using soft masking during training where unaffordable actions are assigned low probability would allow \texttt{CoGA} to slowly catch up to the RL baseline, ultimately lagging in sample efficiency. This limitation is further discussed in Section \ref{sec:limitsfailures}. It is worth noting that the generated affordances can be used in either value-based or policy gradient RL.

\section{Experiments}
\label{sec:experiments}
\subsection{MiniWoB++} MiniWoB++ \citep{pmlr-v70-shi17a, wgeliu2018reinforcementlearningwebinterfaces} consists of a collection of web-based graphical user interface (GUI) tasks, where the goal is to complete tasks by interacting with a simulated webpage. The tasks vary in complexity, ranging from simple actions like clicking a button, to more complex ones like completing a form, or navigating through a series of web elements. Each task is defined by an HTML structure, and the agent's observation consists of a rendered screenshot of the webpage. We use the MiniWob++ environment and action space defined in  \citet{shaw2023pixels}. The action space consists of action types (e.g., \texttt{click}, \texttt{begin\_drag}) and $(x,y)$ pixel coordinates. The affordances are on both action types and pixel coordinates. Every pixel observation is 160x210 pixels, which we divide into 32 bins as in \citet{shaw2023pixels}. We discard text-entry tasks, and therefore the \texttt{type} and keyboard actions. Concretely, this results in an action space of 4x1024. 

The rewards are defined in $(-1,1)$. Positive rewards (success of 1) are assigned \textit{only} upon successfully completing the task, and negative rewards (success of 0) are assigned otherwise (i.e., sparse rewards). Similar to previous works, to encourage the agent to complete the task faster, we discount the positive rewards by the number of steps taken to complete the task.

\subsection{Methods.}

\paragraph{Reinforcement Learning (RL) Agent.} We use a DQN agent that is built on a convolutional neural network (CNN) backbone for encoding the pixel observations. Additionally, for every observation, we encode the task instruction using Sentence-BERT (SBERT) \citep{reimers2019sentencebertsentenceembeddingsusing} as shown in Figure \ref{fig:coga} (left). Although our method is applicable to both value- and policy-based algorithms, we choose a DQN agent due to improved stability and learning efficiency observed in initial experiments. To further enhance learning stability and efficiency, we specifically use a double DQN \citep{vanhasselt2015deepreinforcementlearningdouble} and prioritized experience replay \citep{schaul2016prioritizedexperiencereplay}. All the following baselines use the same architecture as the RL agent. See Appendix \ref{app:hyperparameters} for hyperparameter details.

\paragraph{\texttt{CoGA}. } During training and inference, the generated affordance script is queried. The returned set of affordable actions are used to mask the action space. The agent can only sample from the affordable actions. Note that we also apply the affordance masks during bootstrapping from $o_{t+1}$, where $o$ denotes the observation.

\paragraph{Behavioral Cloning (BC) Agent. } Due to limited resources and closed-source expert demonstrations performed on the environment we used, we perform behavioral cloning on expert demonstrations that we collected using rollouts from the Pix2Act model \citep{shaw2023pixels}. We filter the trajectories that achieved a reward of less than 0.8.

It should be noted that while many state-of-the-art prior works (e.g., \citep{shaw2023pixels}) on MiniWob++ use large amounts of expert demonstrations, our work focuses on leveraging pre-trained foundation models particularly in the low-data regime with no or limited access to expert demonstrations. Hence, the comparison with BC is limited to scenarios that only use a few expert demonstrations.

\subsection{Results}
\begin{figure*}[ht]
    \centering
    \raisebox{-.5\height}{%
        \begin{minipage}[c]{0.40\linewidth}
            \includegraphics[width=\linewidth]{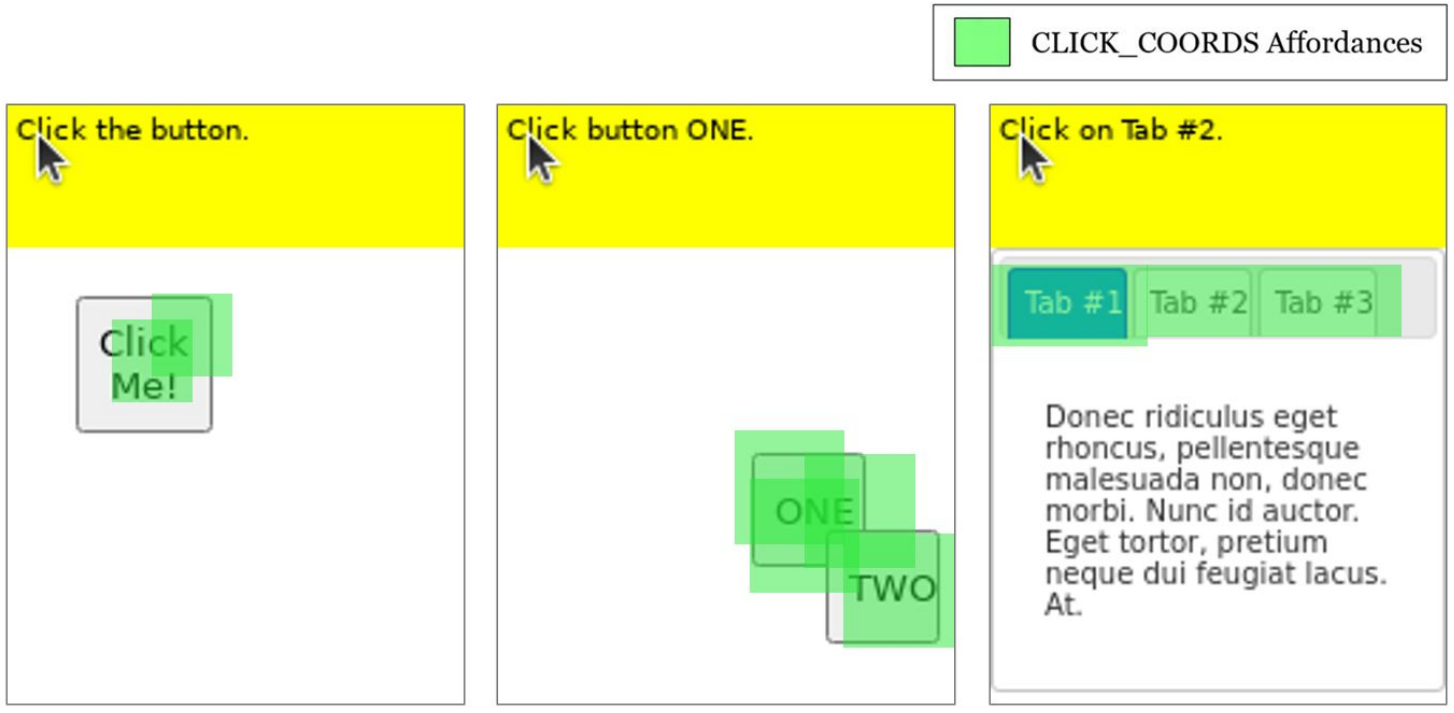}
        \end{minipage}%
    }
    \quad
    \raisebox{-.5\height}{%
        \begin{minipage}[c]{0.40\linewidth}
            \includegraphics[width=\linewidth]{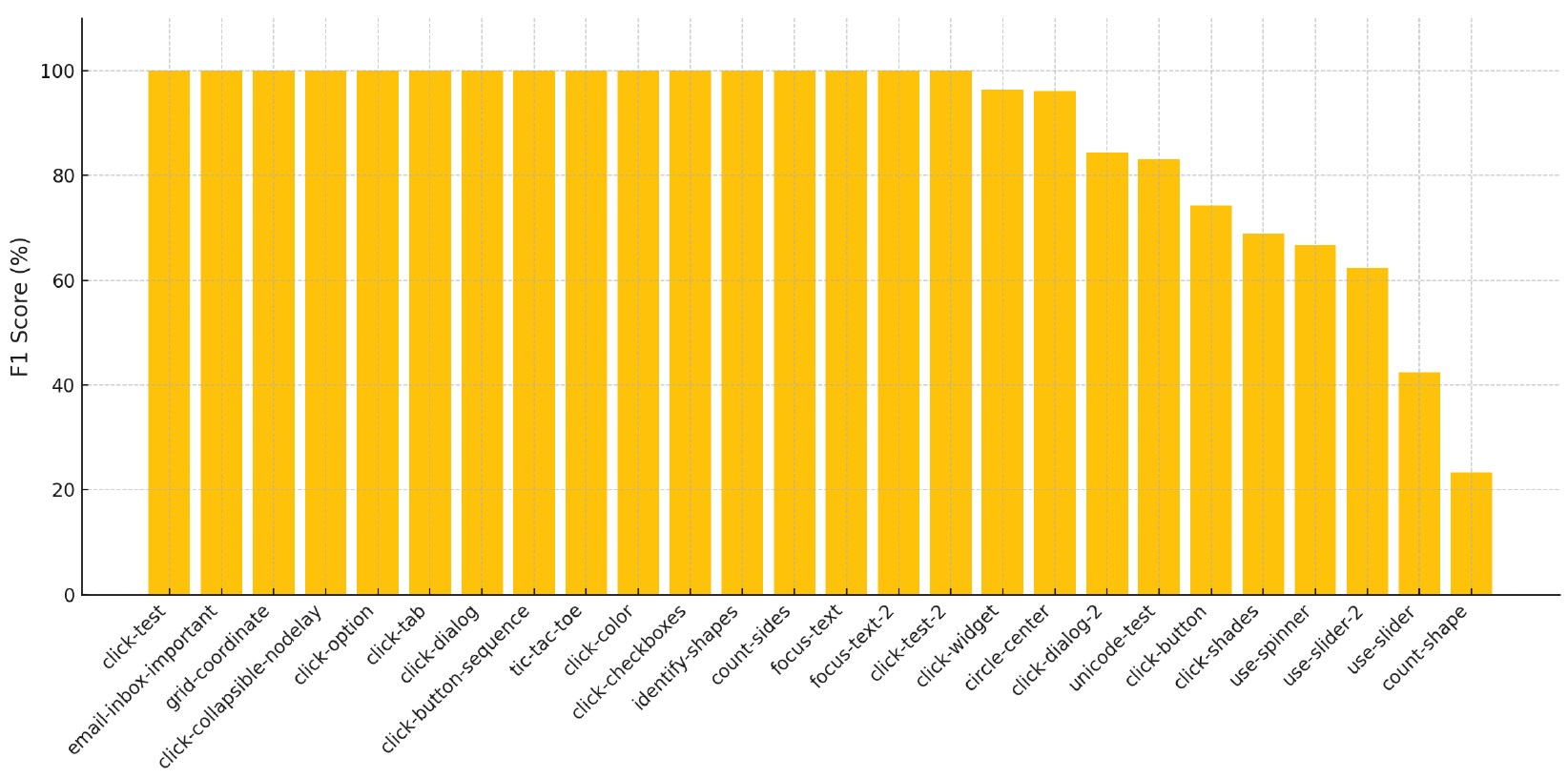}
        \end{minipage}%
    }

    \caption{\textbf{Left:} Examples of returned affordances for three tasks (left to right): \texttt{click-test}, \texttt{click-test-2}, \texttt{click-tab}
    \textbf{Right:} F1-score across tested tasks. We observe that most generated affordance scripts have a high F1-score, implying wide and precise coverage of ground truth affordances.}

    \label{fig:affs}
    
\end{figure*}

\paragraph{\texttt{CoGA}'s Affordance Scripts are Intuitive, (mostly) Accurate, and Precise.}\label{scrtips-qual} We evaluate the quality of the generated affordance scripts qualitatively and quantitatively. Qualitatevely, we observe that the returned affordable actions are intuitive (Figure \ref{fig:affs} left). This is emphasized in instruction-dependent tasks such as \texttt{click-test-2} and \texttt{click-tab} (Figure \ref{fig:affs} left - middle and right). In \texttt{click-test-2} the instruction (e.g., \textit{goal}) is to either click on button \texttt{ONE} or \texttt{TWO}. However, the \textit{intent} is to ``\textit{click a button}", in which case clicking any of the two buttons is affordable, and the policy is learnt over these affordances.

Quantitavely, we measure the quality of the generated affordance scripts using precision and recall, and aggregate them through an F1-score (Figure \ref{fig:affs} right). We define precision as the rate of predicted affordances that have at least one match in the set of ground truth affordances (i.e., identical action type and a corresponding pixel intersection over union (IoU)$>$0). We define recall as the rate of ground truth affordances that have at least one match in the set of predicted affordances. As seen in Figure \ref{fig:affs} (right), most scripts have high recall and precision. For those with low precision but high recall, the affordance set would include more affordable actions than in the ground truth. To this end, in the worst case, \texttt{CoGA} performs on par to the RL agent. See Appendix \ref{app:runsiterationspipeline} for the details of the code generations runs.

\paragraph{\texttt{CoGA} is Orders of Magnitude More Sample Efficient than its RL Agent. }

\begin{figure*}[ht]
    \centering
    \raisebox{-.5\height}{%
        \begin{minipage}[c]{0.4\linewidth}
            \includegraphics[width=\linewidth]{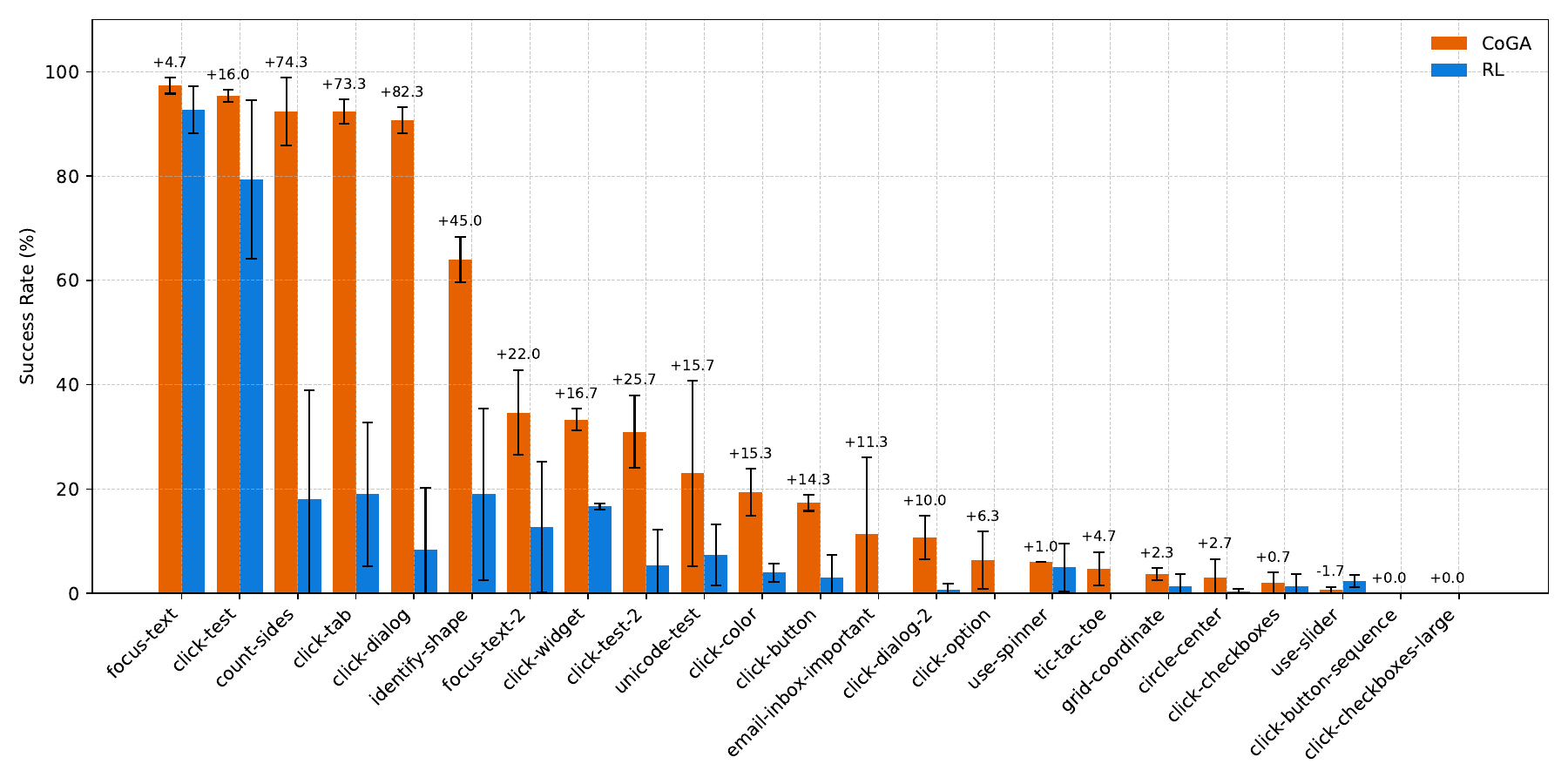}
        \end{minipage}%
    }
    \quad
    \raisebox{-.5\height}{%
        \begin{minipage}[c]{0.50\linewidth}
            \includegraphics[width=\linewidth]{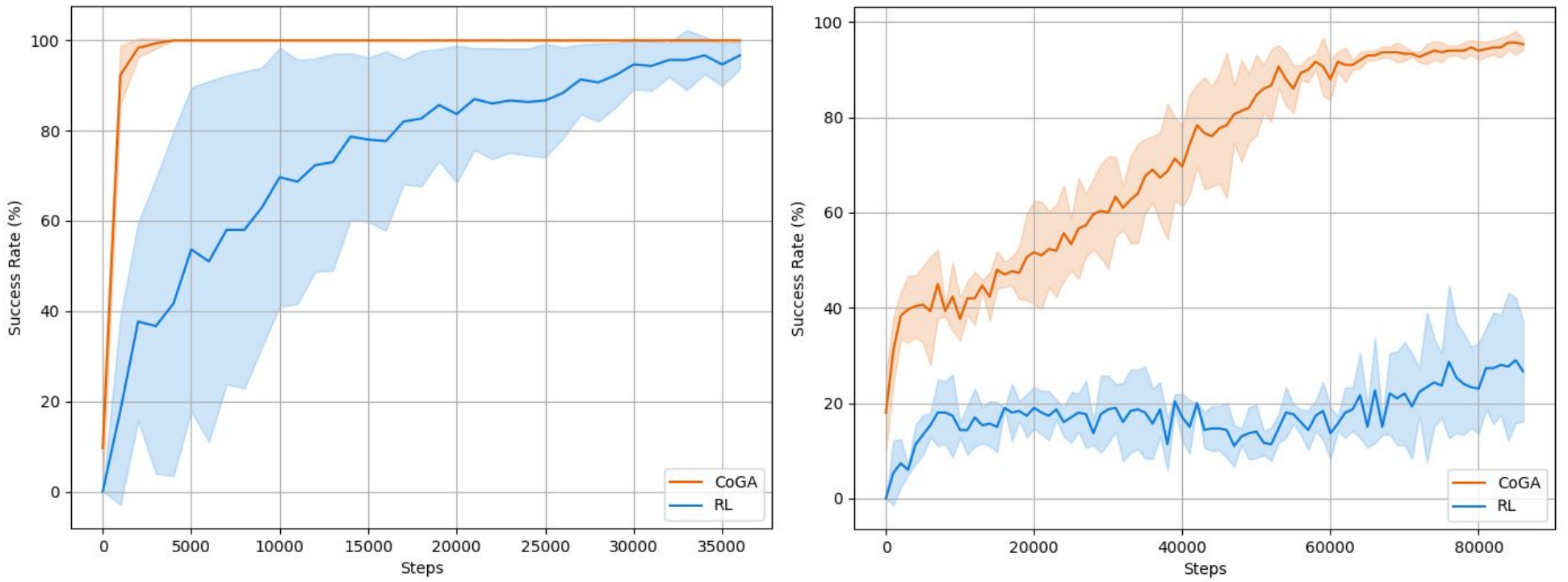}
        \end{minipage}%
    }
    
    \caption{\textbf{Left:} Evaluation success rates at 1000 steps for the  RL agent and \texttt{CoGA} across tasks. We observe that \texttt{CoGA} is over 10 times more sample efficient than the RL agent early in training at only 1000 steps. \textbf{Right:} Evaluation success rate curves for the RL agent and \texttt{CoGA} on \texttt{count-sides} (left) and \texttt{click-test-2} (right)}

    \label{fig:sample-eff}
\end{figure*}

\label{sample-eff}We investigate the effect of constraining the action space using the affordances returned by the generated affordance scripts on the agent's sample efficiency. Following a hyperparameter search (see Appendix \ref{app:hyperparameters}), we report the best evaluation success rates at 1000 steps for the RL agent and \texttt{CoGA} on 23 tasks and over 3 seeds. Due to computational constraints, we evaluate on the tasks which had affordance scripts with a high F1-score, and a few with relatively lower scores to investigate the effect of subpar affordance scripts (e.g., \texttt{use-slider}). As illustrated in Figure \ref{fig:sample-eff} (left), \texttt{CoGA} enables over 10x sample efficiency gains over the RL agent early in training at only 1000 steps, and considerable gains on most tasks when an affordance script has a high F1-score. Sample efficiency curves are also shown in Figure \ref{fig:sample-eff} (right) on \texttt{count-sides} and \texttt{click-test-2} for illustration purposes. 

\paragraph{\texttt{CoGA}'s Affordance Scripts can Generalize within the Same Family of Tasks.}\label{genztion} We define a family of tasks as tasks with the same affordances, but different optimal policies. For instance, \texttt{click-test-2} (Figure \ref{fig:affs} left - middle) and \texttt{click-button-sequence} have an identical GUI. However, in the former, the task is to learn to click on \textit{either} button \texttt{ONE} or \texttt{TWO}, whereas in the latter, the task is to click on button \texttt{ONE} \textit{then} \texttt{TWO}. Thus, we hypothesize that affordances should generalize within the same family of tasks. We compare the best evaluation success rates obtained by using a task's originally generated affordance script (e.g., \texttt{click-button-sequence}) and its relative's generated affordance script (e.g., \texttt{click-test-2}). For reference, we also include the RL agent's performance on the generalization task considered. 

\begin{table}[ht]
\caption{Mean evaluation success rates with standard deviation across 3 seeds. We report the best evaluation success rates on the task using its originally generated affordance script (\texttt{CoGA}-o) and its transfer affordance script (\texttt{CoGA}-t). As a reference, we also include the performance of the RL agent on each task.}
\label{affs-genztion}
\centering
\resizebox{0.7\linewidth}{!}{%
\begin{tabular}{cccc}
\toprule
Task & RL (SR \%) & CoGA-o (SR \%) & CoGA-t (SR \%) \\ \midrule
\texttt{click-button-sequence} & $3.00 \pm 1.00$ & $15.67 \pm 1.15$ & \textbf{$23.67 \pm 1.53$} \\ \midrule
\texttt{focus-text-2} & $80 \pm 28.79$ & \textbf{$100.00 \pm 0.00$} & $100 \pm 0.00$ \\ \midrule
\texttt{click-checkboxes-large} & $0.00 \pm 0.00$ & \textbf{$0.33 \pm 0.58$} & \textbf{$0.67 \pm 0.58$} \\ \bottomrule
\end{tabular}
}\label{tab:generalization}
\end{table}

In Table \ref{tab:generalization}, we observe that a generated affordance script indeed generalizes to its relative tasks. Interestingly, we see that using the generated script of a task's relative can sometimes outperform using a task's original script (e.g., \texttt{click-button-sequence}).  

\begin{figure*}[h!]
    \centering
    \raisebox{-.5\height}{%
      \begin{minipage}[c]{0.5\linewidth}
        \includegraphics[width=\linewidth]{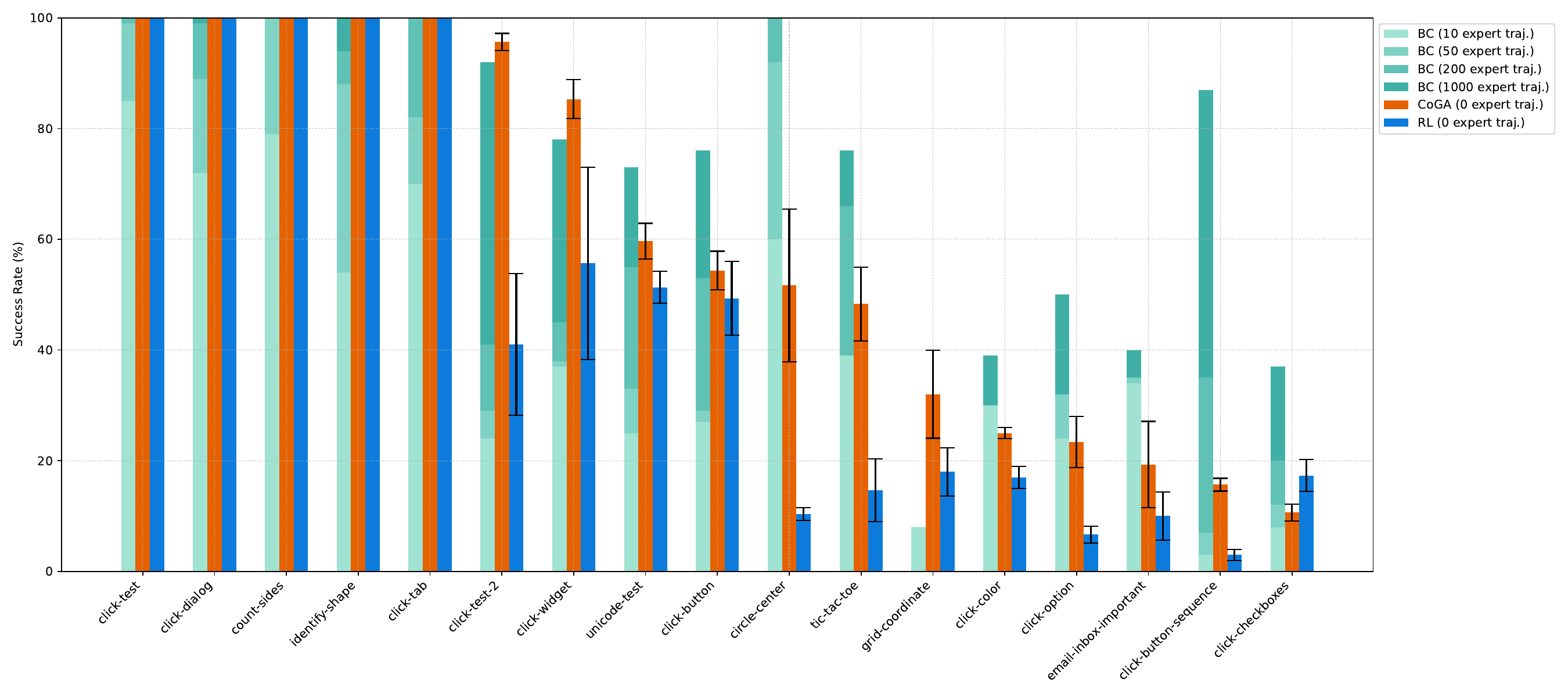}
      \end{minipage}%
    }
    \quad
    \raisebox{-.5\height}{%
      \begin{minipage}[c]{0.4\linewidth}
        \includegraphics[width=\linewidth]{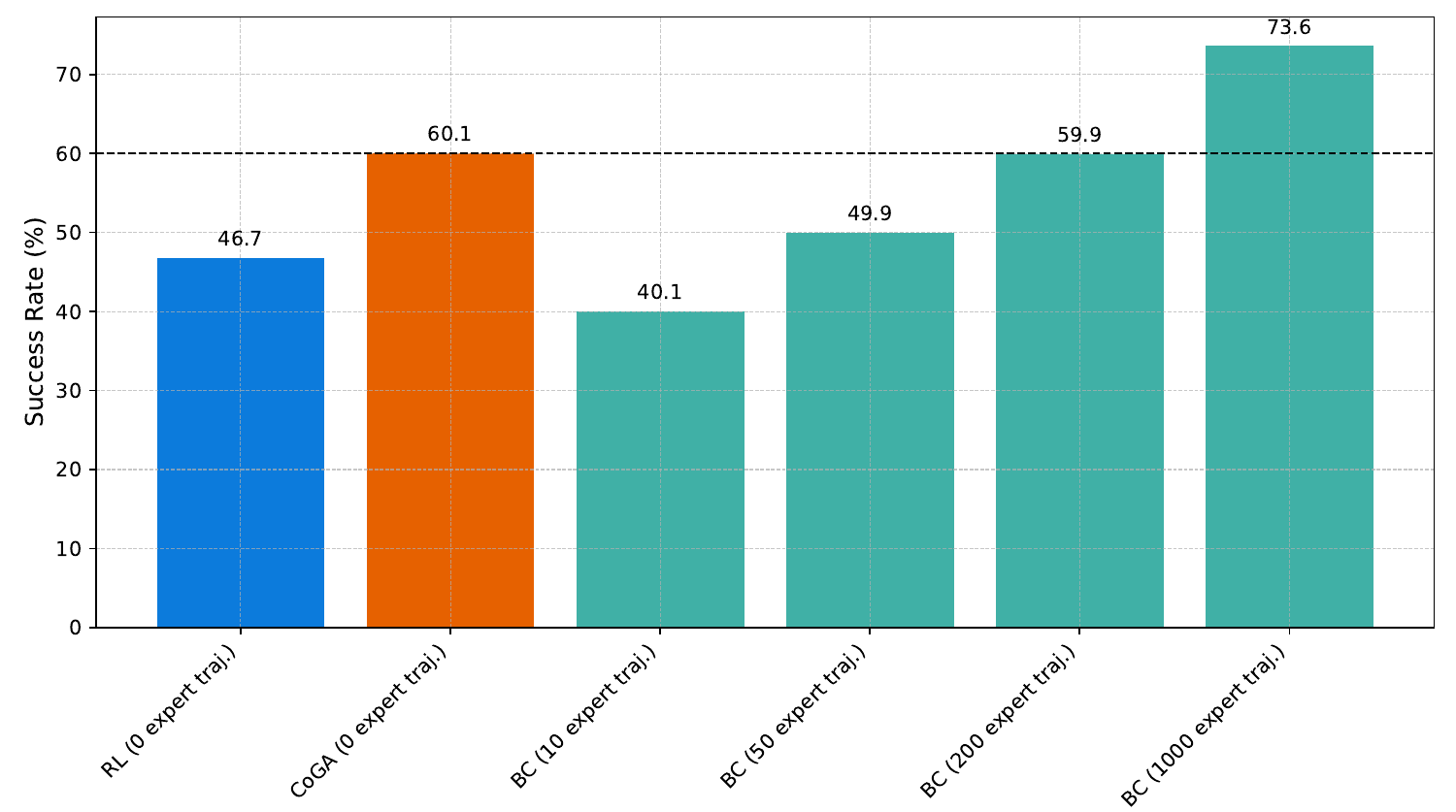}
      \end{minipage}%
    }
    
    \caption{\textbf{Right:} Evaluation success rates across tasks and expert data regimes of the BC agent, the RL agent, and \texttt{CoGA} (mean and standard deviation over 3 seeds). 
    \textbf{Left:} Mean of evaluation success rates across tasks for the RL agent, \texttt{CoGA}, and increasing expert data regimes of the BC agent.}
    
    \label{fig:bc-v-coga}
\end{figure*}

\paragraph{\texttt{CoGA} Performs Better or On Par Compared to the Behavioral Cloning Agent when a Limited Number of Expert Demonstrations are Available.}
\label{coga-bc-vanilla} We consider the low-data regime where a limited number of expert demonstrations are available. With expert demonstrations available, a natural baseline to consider is behavioral cloning. We thus evaluate a BC agent's performance across data regimes (namely, 10, 50, 200, and 1000 expert demonstrations) and compare with the RL agent and \texttt{CoGA} using only self-collected data. We consider the best evaluation success rates for every baseline across a hyperparameter search (see Appendix \ref{app:hyperparameters}) and over a training period of up to 100,000 steps for the RL agent and \texttt{CoGA}, and 30 epochs for the BC agent. As shown in Figure \ref{fig:bc-v-coga} (right), \texttt{CoGA} performs better on average than the BC baseline with up to 200 expert trajectories, beyond which the BC baseline outperforms. However, the RL baseline only outperforms the BC baseline trained on 10 expert demonstrations. Note that we only consider the tasks on which we were able to collect expert trajectories from using the Pix2Act model. These results demonstrate the impact of constraining the action space using affordances on an agent's performance. Particularly, by combining a limited number of expert data and \texttt{CoGA}, one could expect significant boosts in performance that potentially match that of a BC+RL agent using higher expert data regimes.

\section{Related Work}
\label{sec:relatedwork}
\paragraph{MiniWob++.} MiniWoB was originally introduced by \citet{pmlr-v70-shi17a} and extended to MiniWob++ \citep{wgeliu2018reinforcementlearningwebinterfaces} through additional tasks. Prior works have investigated visual- and structure-based approaches. \citet{pmlr-v70-shi17a} investigate DOM and pixel observation with an average of 200 human demonstrations per task to train a BC policy followed by RL. CC-Net \citep{humphreys2022datadrivenapproachlearningcontrol} extends this approach by increasing the amount of expert demonstrations to a total of 2.4 million and scaling the model architecture. Due to the limited availability of DOM elements, Pix2Act \citep{shaw2023pixels} considers pixel-only observations. It uses a transformer base model that was pre-trained to map screenshots to HTML structures. They first fine-tune their model using BC on approximately 1 million expert demonstrations, followed by Monte Carlo Tree Search for policy improvement. More recently, \citet{cheng2024seeclickharnessingguigrounding} employ pre-training to ground large VLMs to GUI coordinates and use them as the GUI agent. Other approaches have considered language-based methods through webpage structural information. \citet{wgeliu2018reinforcementlearningwebinterfaces} propose workflow-guided exploration (WGE) by using expert demonstrations to learn high-level ``workflows" through which the agent learns to select appropriate actions within the workflow through RL. Although similar to \texttt{CoGA} in constraining the action space and minimizing the amount of expert demonstrations required, WGE relies on DOM inputs and parametric learning of the workflows. In contrast, \citet{kim2023languagemodelssolvecomputer} employ a large language model (LLM) as the agent using HTML code as input. Motivated by contemporary approaches to train an agent to interact with a GUI similarly to humans and without relying on OS- or web-specific APIs \citep{cua2025,anthropicCUA}, our method seeks to explore the \textit{low expert data regime} using \textit{pixel-only observations} and a unified pixel and action type action space. Considering the significant computational and financial costs that VLM agents incur, we argue that only selected prior knowledge of the VLM is relevant for a web agent. Therefore, we propose to distill the relevant information from the VLM through code as affordances.

\paragraph{VLMs for Affordances.} VLMs have been used to infer affordances by leveraging their ability to reason about visual and textual information. Using a visual question-answering prompting technique,  MOKA \citep{liu2024mokaopenworldroboticmanipulation} and  KAGI \citep{lee2024affordanceguidedreinforcementlearningvisual} employ VLMs to predict keypoint and waypoint affordances from pre-marked visual observations to guide open-world robotics. VoxPoser \citep{huang2023voxposercomposable3dvalue} uses LLMs to infer affordances for open-world robotic manipulation given free-form language. They leverage LLMs to write code that interacts with a VLM to compose 3D value maps for grounding the agent's observation space. Our method is complementary to previous work. As detailed in Sec. \ref{sec:coga}, we build upon their successes to infer affordances using the VLM through pre-marked observations and code.

\paragraph{Code Generation for Reinforcement Learning.} Recent works have investigated the role of code generation by foundation models (e.g., LLMs, VLMs) as a mechanism for improving RL sample efficiency. Code as Policies \citep{liang2023code} introduced a paradigm to prompt LLMs to generate structured programs that serve as policies for robots. Although they do not use VLMs directly to generate code due to their rather limited capabilities at the time, they leverage them for object detection, localization, and segmentation through the generated code. Code as Reward \citep{venuto2024code} proposes using VLMs to produce dense reward functions as code, improving sample efficiency in RL. Similarly, EUREKA \citep{ma2024eurekahumanlevelrewarddesign} is a framework for LLMs to generate and iteratively improve reward functions as code, outperforming human-engineered rewards. Octopus \citep{yang2024octopusembodiedvisionlanguageprogrammer} uses a VLM to generate code for planning and manipulation. Likewise, Voyager \citep{wang2023voyager} leverages LLMs to generate code as actions. VoxPoser \citep{huang2023voxposercomposable3dvalue} enables LLMs to write code that interacts with a VLM and ground the agent's observation space through the inferred affordances. We build on the insights of previous work in this regard, particularly on that of \citet{venuto2024code}.

\section{Discussion}

\subsection{Limitations and Failure Modes} 
\label{sec:limitsfailures}
\paragraph{1) VLM Pixel Mapping.} VLMs struggle to parse an image into pixel coordinates \citep{cheng2024seeclickharnessingguigrounding}. To mitigate this issue, we superimposed a granular coordinate system onto an observation when showing it to the VLM for template image extraction. However, the VLM struggles in specifying exact correct coordinates, which affects the quality of the template images. Consequently, this can decrease the recall and precision of the predicted affordances. 

\paragraph{2) Template Matching.} Template matching is limited to detecting isomorphic objects across observations. Although this can be mitigated by lowering the template matching threshold (which the critique VLM self-iterates on), it is particularly a limitation for varying text-based objects. To this end, we have experimented with optical character recognition (OCR) tools (e.g., \texttt{pytesseract}). However, due to inconsistent OCR, we decided to exclude most tasks with varying text-based observations from our work. In future work, one might explore using more robust OCR tools. Additionally, in some tasks like \texttt{bisect-angle} where the agent must click on the location that bisects an angle in half, the VLM fails to extract valid template images as it is unclear how to express affordances in this way. 

\paragraph{3) VLM Code Generation.} The recall and precision of the predicted affordances depend on the correctness of the generated code. Given perfect templates but incorrect reasoning about the affordable action types and their pixel coordinates, affordances would be compromised. We mitigated this issue through sequential chain-of-thought prompting \citep{wei2023chainofthoughtpromptingelicitsreasoning} and by using the critique VLM. However, this limitation could be more prevalent in more complex tasks where the script must pick the relevant subset of intents and their corresponding affordances from a set of intents in different situations (e.g., \texttt{email-inbox}). Finally, an important limitation was the amount of time taken to generate code for tasks that contained a larger number of affordable elements (e.g., \texttt{drag-shapes}).

\paragraph{4) VLM Code Verification.} We had to manually label a set of 5 test cases to evaluate the quality of the generated code. Although it is typical to manually create test cases for software systems, one could investigate the VLM's ability to create its own test cases. It is also important to note that the critique VLM is not always correct. For instance, it could label a code as ``failed'' while scoring high recall and precision. This could be potentially mitigated by incorporating a self-improving paradigm to the critique VLM, whereby it could learn from its mislabeled evaluations of the code based on the results from the  manual test cases. Finally, an important limitation that we faced was token limits, which prevented further iterations of the code based on the critique VLM's feedback. This was the case for tasks like \texttt{click-shades}, \texttt{count-shape}, and \texttt{use-slider-2}.

\subsection{Future Work}
\label{sec:discfuture}
Given perfect affordances, \texttt{CoGA}'s performance is limited to the strength of the backbone RL agent (e.g., network architecture, RL algorithm, hyperparameter sweeps), particularly in tasks that require sequential decision-making over multiple steps and are partially observable. We especially note this in tasks like \texttt{click-checkboxes} and \texttt{click-checkboxes-large} for example. A promising direction for future work is to augment more competent RL agents with \texttt{CoGA}, as our method is complimentary to any RL algorithm.

\section*{Acknowledgements}
LC and DV gratefully acknowledge funding from the FRQNT Master's and Doctoral Training Scholarships, and FK from the FRQS Master's Training Scholarship. The authors would like to thank Mohammad Sami Nur Islam for helping set up and run experiments in early iterations of the project. The authors are grateful for Mohammad Sami Nur Islam, Yash More, and Nikhil Vemgal for their computational resources, and Bernardo Avila Pires for providing valuable feedback on an early version of the draft. 

\bibliography{iclr2025_conference}
\bibliographystyle{iclr2025_conference}

\newpage
\appendix

\appendix
\onecolumn
\section{Prompting Pipeline for Generating Intents and the Initial Affordance Script}
\label{appendix:promptingpipeline}

\begin{itemize}
    \item This is an image of a web environment. The agent is the cursor. It can click anywhere on the screen. There may also be other relevant objects that the agent can interact with. The task is to \{task description\}, but note that the task's specific utterances can vary. Here are some examples: \{example utterances\}. Can you give a list of the most important elements in this image, ensuring it applies to all the utterances above, not just the specific instance?  Give me a list of elements and concise names with description. Do not include the background.
    \item Based on your description of these elements in the environment, what do you think the agent affords in this environment? The final goal completion and reading the instruction are not affordances. This list should be concise and only contain affordances that are actionable directly by the agent. If multiple affordances can be combined, please combine them into one modular affordance. Please only return a python list of affordance names.
    \item For each affordance: 
    \begin{itemize}
        \item what are the most relevant objects that you need to identify in this environment to check if the affordance is possible? Give me the minimum list concisely but precisely. Do not give a generic answer such as `shapes'. Please end your answer by returning a python list of object names. Ensure that the object name does not contain any spaces. The cursor should not be in the list.
    \end{itemize}
    \item For each object \{obj\}, for each gridded image:
    \begin{itemize}
        \item Explain to a 5 year old step by step how to visually identify a \{obj\} in such an image.
        \item Look at the image grid and find the \{obj\}. Return the bounding box coordinates as [x\_left, y\_upper, x\_left+width, y\_upper+height] where width and height describe the size of the box. Ensure that 0$\leq$x$\leq$160 and 0$\leq$y$\leq$210. Return only the list.
        \item The template image for an instance of \{obj\} has been saved in \{template\_path\}. You can use this template image for \{obj\} detection using template matching when needed.
    \end{itemize}
    \item Here is a script that can be used for object detection using template matching. It will be referred to as match\_template. You can use it with the template paths, but do not modify it. Here is the script:
    \par\noindent\texttt{def match\_template(image\_array, template\_path, save\_path):}
    \par\noindent\texttt{\ \ \ \ \# The main image is provided as an array}
    \par\noindent\texttt{\ \ \ \ main\_image = image\_array}
    \par\noindent\texttt{\ \ \ \ template\_image\_path = template\_path}
    \par\noindent\texttt{\ \ \ \ \# Load the template image in grayscale}
    \par\noindent\texttt{\ \ \ \ template\_image = cv2.imread(template\_image\_path, cv2.IMREAD\_GRAYSCALE)}
    \par\noindent\texttt{\ \ \ \ if main\_image.ndim == 3 and main\_image.shape[2] == 3:}
    \par\noindent\texttt{\ \ \ \ \ \ \ \ main\_image\_gray = cv2.cvtColor(main\_image, cv2.COLOR\_RGB2GRAY)}
    \par\noindent\texttt{\ \ \ \ else:}
    \par\noindent\texttt{\ \ \ \ \ \ \ \ main\_image\_gray = main\_image}
    \par\noindent\texttt{\ \ \ \ main\_image\_rgb = cv2.cvtColor(main\_image\_gray, cv2.COLOR\_GRAY2RGB)}
    \par\noindent\texttt{\ \ \ \ w, h = template\_image.shape[1], template\_image.shape[0]}
    \par\noindent\texttt{\ \ \ \ result = cv2.matchTemplate(main\_image\_gray, template\_image, cv2.TM\_CCOEFF\_NORMED)}
    \par\noindent\texttt{\ \ \ \ threshold = 0.5}
    \par\noindent\texttt{\ \ \ \ locations = np.where(result >= threshold)}
    \par\noindent\texttt{\ \ \ \ bounding\_boxes = []}
    \par\noindent\texttt{\ \ \ \ for pt in zip(*locations[::-1]):}
    \par\noindent\texttt{\ \ \ \ \ \ \ \ bounding\_box = (pt[0], pt[1], pt[0] + w, pt[1] + h)}
    \par\noindent\texttt{\ \ \ \ \ \ \ \ bounding\_boxes.append(bounding\_box)}
    \par\noindent\texttt{\ \ \ \ \ \ \ \ cv2.rectangle(main\_image\_rgb, (pt[0], pt[1]), (pt[0] + w, pt[1] + h), (0, 255, 0), 2)}

    \par\noindent\texttt{\ \ \ \ result\_image\_path = f`matched\_templates/{save\_path}'}
    \par\noindent\texttt{\ \ \ \ cv2.imwrite(result\_image\_path, main\_image\_rgb)}
    \par\noindent\texttt{\ \ \ \ print(f"Number of matches found: \{len(bounding\_boxes)\}")}
    \par\noindent\texttt{\ \ \ \ return bounding\_boxes}

    \item Write a step-by-step strategy for determining which affordance to use at a given state of the environment. If multiple affordances apply, your strategy should return them in hierarchical order. Please do not write any code yet. Reading the instruction should not be part of the strategy. The strategy should be independent of the instruction. Make sure your strategy is as specific as possible and that it can generalize to all possible states that the agent can be in.
    \item For each affordance:
    \begin{itemize}
        \item Write a step-by-step strategy for determining which action(s) is/are affordable at a given state for the intent of \{aff\} by examining a current state image. You should return the set of affordable actions to complete the intent of \{aff\}.  Please do not write any code yet. Make sure your strategy is as specific as possible. When deciding on actions, note that CLICK\_COORDS and DBLCLICK\_COORDS automatically move the cursor and perform the click action, so there is no need to separately use MOVE\_COORDS before them The possible actions are: \{action\_set\}. This list contains both action names and their textual descriptions. You should refer to actions by their names in all responses. The PRESS\_KEY action type issues a key combination. Each key combination in the allowed\_keys config follow the rules: Modifiers are specified using prefixes `C-' (Control), `S-' (Shift), `A-' (Alternate), or `M-' (Meta). Printable character keys (a, 1, etc.) are specified directly. Shifted characters (A, !, etc.) are equivalent to `S-' + non-shifted counterpart. Special keys are enclosed in `<…>'. The list of valid names is specified inminiwob.constants.WEBDRIVER\_SPECIAL\_KEYS. Example valid key combinations:`7, `Enter, `C-S-ArrowLeft'. To specify relevant x,y coordinates for a selected action, you can use the template matching script for the relevant objects.
    \end{itemize}
    \item Write a step-by-step strategy for combining the intents and their corresponding set of affordable actions at a given state. After writing your strategy, write an outline of a code using comments. The purpose of the affordable actions returned by the code should not be to solve the task specified in the instruction, but to just help give a prior to an RL agent to learn how to solve the task by helping it narrow down its exploration space. You should return a set of affordable actions to complete the selected intent. This script should not require any other input than the image.
    \item Write code to implement your strategy for selecting the required affordance and its corresponding set of affordable actions at a given state. First write an outline of the code with comments, then write the filled in code. The purpose of the affordable actions returned by the code should not be to solve the task specified in the instruction, but to just help give a prior to an RL agent to learn how to solve the task by helping it narrow down its exploration space, wihtout telling it the exact solution. You can use the object ID scripts and match\_template if needed, but do not modify them. If you use the match\_template, make sure to use all of the templates inthe required template's path. The template paths available are: \{template\_paths\}. Make sure your implementation can generalize to all possible states that the agent can be in and pay close attention to what you have written in the steps. You should return a set of affordable actions to complete the selected affordance. This script should not require any other input than the image. The script should be fully executable as is, and should not have placeholder values or pseudocode. It should also be robust to any detection imperfections. Make sure that the coordinate affordances returned are those of the bounding box around the relevant object in the form of [x\_left,y\_up,x\_right,y\_down]. The returned affordances should be a list of dictionaries with the format {{`action': action\_name, `coords': [x\_left,y\_up,x\_right,y\_down]}}. x and y should not be the centers. The action\_name should be exactly as specified in the list of actions. The function that returns the affordable actions must be called `determine\_affordable\_actions(image)', where the image is a numpy array. Do not change the function's signature, use it as is. The template paths to use must be instantiated inside the function. Ony return the code. DO NOT INCLUDE AN EXAMPLE USAGE.
\end{itemize}

\section{Prompting Pipeline for Code Review \& Regeneration}
\label{app:verifpipeline}
\begin{itemize}
    \item For up to 3 times, terminating early if the code is verified as correct:
    \begin{itemize}
        \item \textit{(To critique VLM)} This is an image of a web environment. The agent is the cursor. It can click anywhere on the screen. There may also be other relevant objects that the agent can interact with. The task is to \{task\_description\}, but note that the task's specific utterances can vary. Here are some examples: \{example\_utterances\}. Do not output anything yet.
        \item \textit{The code is executed on 2 example observations for error handling. If the code runs smoothly:}
        \begin{itemize}
            \item \textit{(To critique VLM)} Please review this code and find problems with it, if any. The affordable actions returned by this code on the first image are: \{actions\_1\}. The affordable actions returned by this code on the second image are: \{actions\_2\}. The purpose of the affordable actions returned by the code should not be to solve the task specified in the instruction, but to just help give a prior to an RL agent to learn how to solve the task by helping it narrow down its exploration space, without telling it the exact solution. Please think about whether these returned affordable actions are suitable for each test case image provided. If the affordable actions are suitable, please only return a string {{Status: Pass, Reasoning: step-by-step reasoning for why the actions are correct}}. Else, only return {{Status: Fail, Reasoning: step-by-step reasoning for why the actions are incorrect, Critique: Point-based feedback on how to solve the issues found in the code}} (MUST NOT BE JSON). Do not modify the code yet.
        \end{itemize}
        \item \textit{However, if there is an error in the code:} 
        \begin{itemize}
            \item \textit{(To critique VLM)} Please review this code as it is throwing this error: \{error\_trace\}. Only return \{Status: Fail, Reasoning: step-by-step reasoning for why the code is failing, Critique: Point-based feedback on how to solve the issues found in the code\} (MUST NOT BE JSON). Do not modify the code yet. The determine\_affordable\_actions function must only take the image as argument.
        \end{itemize}
        \item \textit{(To critique VLM)} Based on the issues you have found in the code, please improve your code. You should only rewrite the code. Do not mention anything else in your answer. DO NOT INCLUDE EXAMPLE USAGE.
    \end{itemize}
\end{itemize}

\section{RL and \texttt{CoGA} Hyperparameter Search Details}
\label{app:hyperparameters}
The hyperparameters used consistently across all tasks are summarized in Table~\ref{tab:hyperparams}.

\begin{table}[h]
    \centering
    \caption{Hyperparameters used across tasks.}
    \begin{tabular}{|c|c|}
        \hline
        \textbf{Hyperparameter} & \textbf{Value} \\
        \hline
        Learning rate ($lr$) & $1\times10^{-5}$ \\
        Gradient Norm Clipping threshold ($clip$) & $1.0$ \\
        Epsilon decay ($eps\_decay$) & $5000$ \\
        Batch size ($batch\_size$) & $64$ \\
        Initial epsilon ($eps\_start$) & $0.6$ \\
        Replay buffer size ($buffer\_size$) & $40000$ \\
        Number of timesteps per update ($n\_timesteps$) & $50$ \\
        Discount factor ($\gamma$) & $0.9$ \\
        Target network update parameter ($\tau$) & $\{1\times10^{-5}, 1\times10^{-6}, 1\times10^{-7}\}$ \\
        \hline
    \end{tabular}
    \label{tab:hyperparams}
\end{table}

For tasks that did not reach a success rate of 100\%, an additional hyperparameter search was performed. This focused on increasing the batch size to $batch\_size=\{128, 256\}$ with $\tau = \{1\times10^{-5},1\times10^{-6}\}$, $buffer\_size=10000$, $eps\_decay=10000$, $eps\_start=0.3$.

We used a prioritized replay buffer with parameters $\alpha = 0.6 $ and  $\beta = 0.4$. All agents use a CNN  with five convolutional layers (32, 64, 128, 256, 512 filters, kernel size 3x3, stride 1, padding 1), each followed by batch normalization and ReLU activation, with max-pooling (2x2, stride 2) after the second, fourth, and fifth layers. The extracted features are flattened and passed through two fully connected layers (1024, 384 neurons, ReLU activations), of which the first fully connected layer's features are concatenated with the SBERT instruction embeddings. The model outputs Q-values for 4 action types and 1024 discretized pixel coordinates (32 x 32 bins). We report results over seeds $\{0,1,2\}$ and use an AdamW optimizer with default parameters and weight decay of $10^{-5}$.

\section{BC Hyperparameter Details}
\label{app:BC_params}
The hyperparameters used for the BC agent are summarized in Table~\ref{tab:BC_params}.
\begin{table}[ht]
    \centering
    \caption{Hyperparameters used for the BC agent.}
    \begin{tabular}{|c|c|}
        \hline
        \textbf{Hyperparameter} & \textbf{Value} \\
        \hline
        Learning rate ($lr$) & $1\times10^{-4}$ \\
        Batch size ($batch\_size$) & $32$ \\
        Epochs ($epochs$) & $30$ \\
        \hline
    \end{tabular}
    \label{tab:BC_params}
\end{table}

\section{Example Generated Affordance Scripts \& Templates}
\label{app:scriptstemplates}
We here show 4 examples of generated affordance scripts and corresponding templates. The first 2 are successful examples, and the last 2 are examples of limitations and failure modes (as shown by the F1-scores in Figure \ref{fig:affs} right).
\begin{figure}[ht]
\begin{center}
\centerline{\includegraphics[width=\columnwidth]{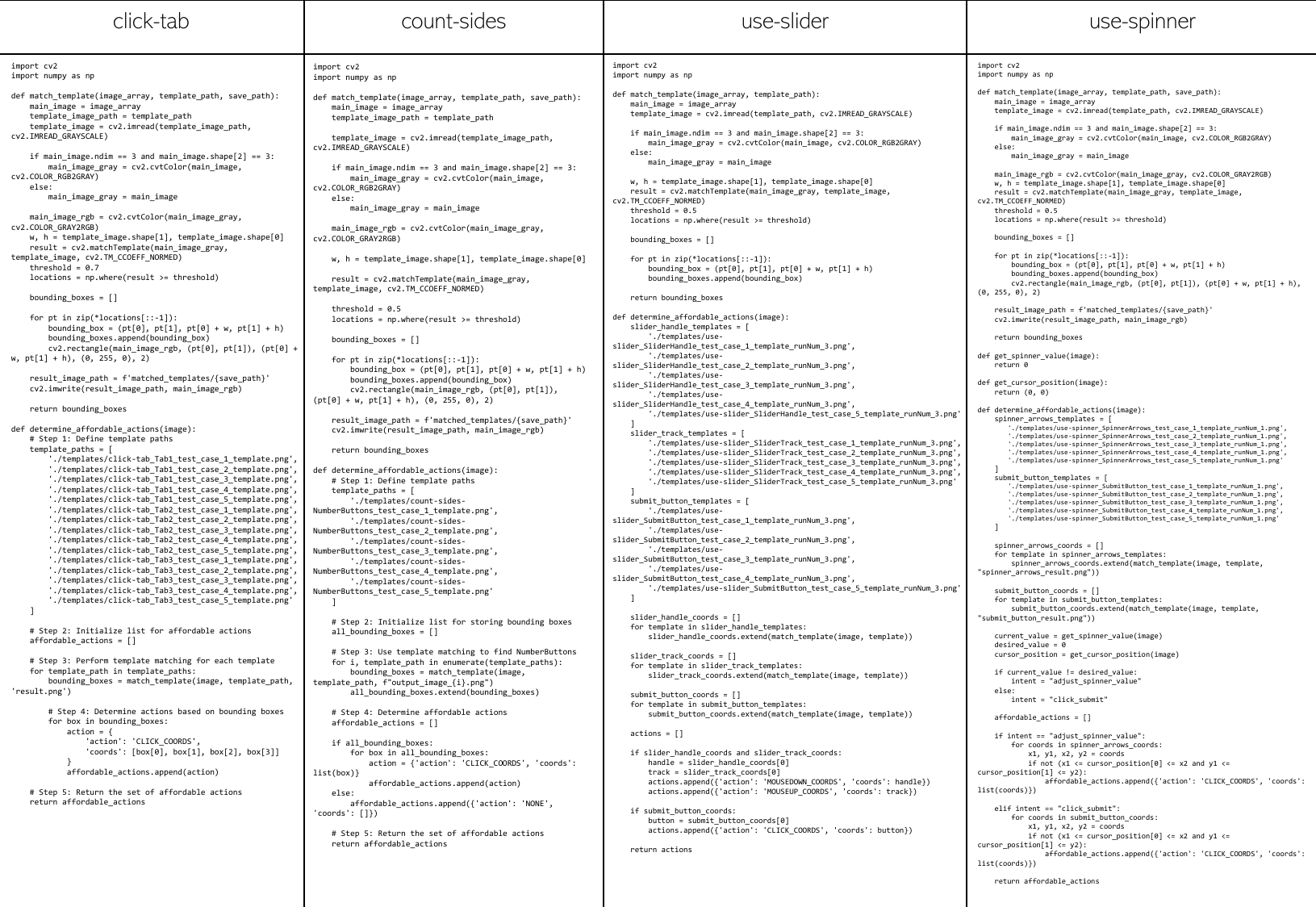}}
\caption{Example scripts across 2 successful tasks (click-tab, count-sides) and 2 unsuccessful tasks (use-slider, use-spinner).}
\end{center}
\end{figure}

\begin{figure}[ht]
\vskip 0in
\begin{center}
\centerline{\includegraphics[width=\columnwidth]{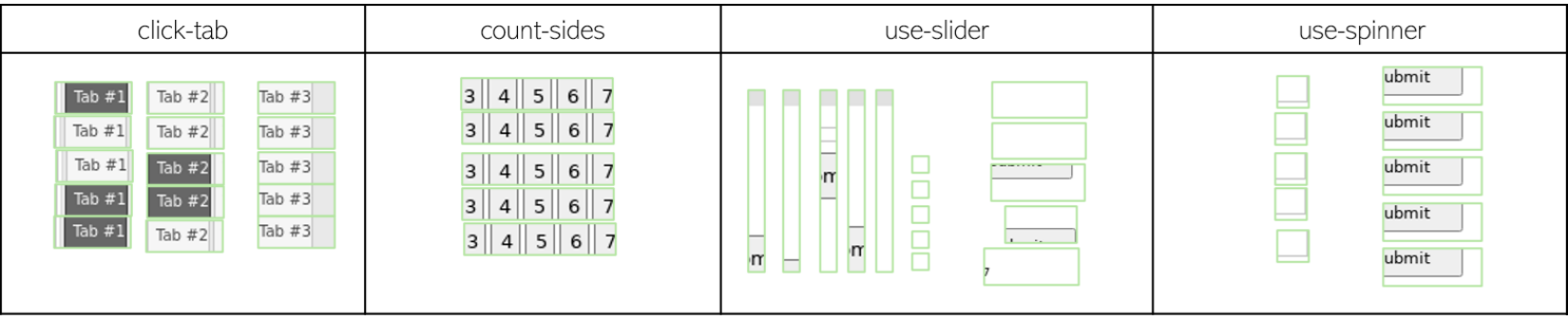}}
\caption{Example template images across 2 successful tasks (click-tab, count-sides) and 2 unsuccessful tasks (use-slider, use-spinner).}
\end{center}
\end{figure}

\section{Number of Runs \& Code Generation Iterations Across Tasks}
\label{app:runsiterationspipeline}
\begin{table}[ht]
    \centering
    \caption{Number of runs and code generation iterations in the best run across tasks. The former represents the total VLM executions per task, while the latter denotes the number of code generation iterations in the most successful execution.}
    \vspace{0.2in}
    \resizebox{\linewidth}{!}{%
    \begin{tabular}{|l|c|c|}
        \hline
        \textbf{Task} & \textbf{Number of VLM Runs} & \textbf{Number of Code Generation Iterations in the Best Run} \\
        \hline
        Click-test & 1 & 1 \\
        Click-tab & 2 & 2 \\
        Circle-center & 1 & 1 \\
        Click-test-2 & 1 & 2 \\
        Focus-text-2 & 1 & 2 \\
        Focus-text & 1 & 2 \\
        Count-sides & 1 & 1 \\
        Identify-shapes & 1 & 1 \\
        Click-checkboxes & 1 & 1 \\
        Click-color & 2 & 1 \\
        Tic-tac-toe & 3 & 2 \\
        Click-button-sequence & 1 & 2 \\
        Click-dialog & 2 & 2 \\
        Click-dialog-2 & 1 & 3 \\
        Click-option & 1 & 2 \\
        Unicode-test & 3 & 2 \\
        Click-button & 3 & 2 \\
        Click-widget & 3 & 2 \\
        Email-inbox-important & 3 & 3 \\
        Grid-coordinate & 2 & 1 \\
        Click-collapsible-nodelay & 2 & 1 \\
        Click-shades & 3 & 1 \\
        Count-shape & 3 & 2 \\
        Use-slider-2 & 3 & 1 \\
        Use-slider & 3 & 3 \\
        Use-spinner & 3 & 3 \\
        \hline
    
    \end{tabular}%
    }
    \label{tab:task_data}
\end{table}

%%%%%%%%%%%%%%%%%%%%%%%%%%%%%%%%%%%%%%%%%%%%%%%%%%%%%%%%%%%%%%%%%%%%%%%%%%%%%%%
%%%%%%%%%%%%%%%%%%%%%%%%%%%%%%%%%%%%%%%%%%%%%%%%%%%%%%%%%%%%%%%%%%%%%%%%%%%%%%%

\end{document}